%% file: main_arxiv.tex
\documentclass[runningheads]{llncs}

\usepackage{eccv}

\usepackage{eccvabbrv}

\usepackage{graphicx}
\usepackage{booktabs}
\usepackage{multirow}
\usepackage{caption}
\usepackage[title,toc,titletoc,header]{appendix}
\usepackage[accsupp]{axessibility}  

\newcommand{\name}{LMM\xspace}
\newcommand{\dname}{MotionVerse\xspace}
\newcommand{\attname}{ArtAttention\xspace}

\usepackage[pagebackref=true,breaklinks=true,letterpaper=true,colorlinks,bookmarks=false]{hyperref}
\newcommand{\repeatthanks}{\textsuperscript{\thefootnote}}

\usepackage{orcidlink}

\begin{document}

\title{Large Motion Model for Unified Multi-Modal Motion Generation} 


\author{Mingyuan Zhang\thanks{co-first authors; $^{\dagger}$ corresponding author}$^{,1}$,
Daisheng Jin\repeatthanks$^{,1}$,
Chenyang Gu\repeatthanks$^{,1}$,
Fangzhou Hong$^{1}$,
Zhongang Cai$^{1,2}$,
Jingfang Huang$^{1}$,
Chongzhi Zhang$^{1}$,
Xinying Guo$^{1}$, \\
Lei Yang$^{2}$,
Ying He$^{1}$,
Ziwei Liu$^{\dagger,1}$
}

\institute{$^{1}$S-Lab, Nanyang Technological University, $^{2}$SenseTime Research}

\authorrunning{Zhang, Jin, Gu et al.}

\institute{S-Lab, Nanyang Technological University, Singapore \and
SenseTime Research, China \\
\vspace{8pt}
Project Page: \url{https://mingyuan-zhang.github.io/projects/LMM.html}
}

\maketitle

\input{figures/teaser}
\input{secs/0_abstract}
\input{secs/1_intro}
\input{secs/2_related}
\input{secs/3_method}
\input{secs/4_experiment}

\input{secs/5_conclusion}
\newpage

\bibliographystyle{splncs04}
\bibliography{main}
\newpage

\begin{subappendices}
\renewcommand{\thesection}{~\Alph{section}}
\input{secs/6_supp}

\end{subappendices}
\end{document}

%% file: figures/teaser.tex
\begin{figure}[h]
    \vspace{-20pt}
    \centering
    \includegraphics[width=\linewidth]{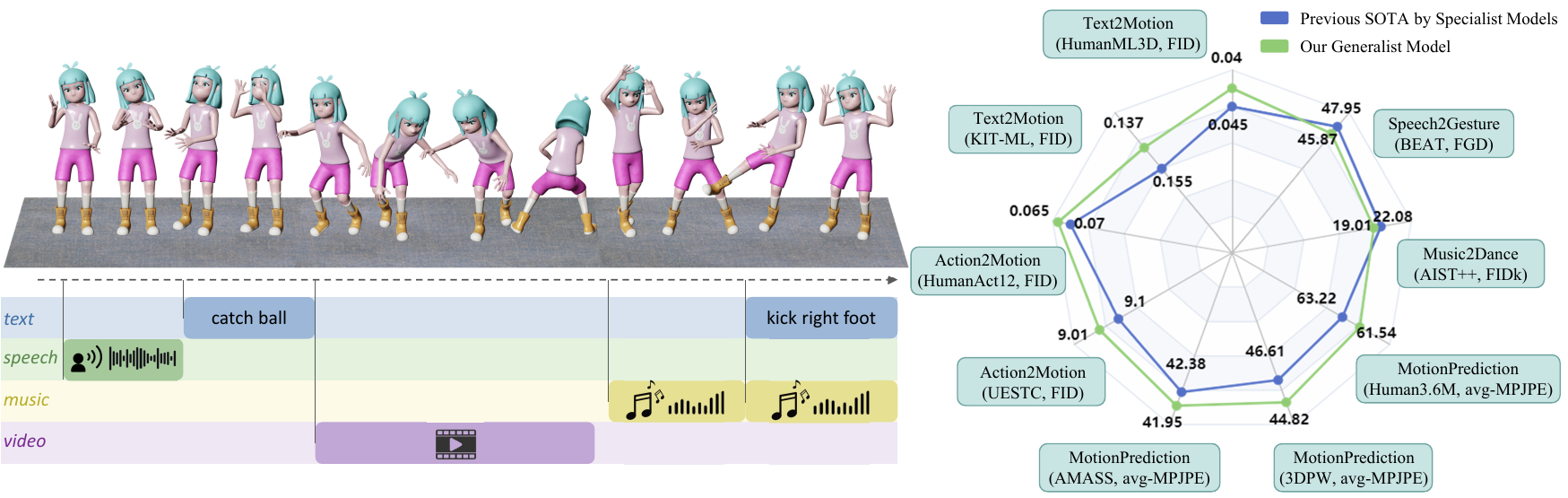}
    \caption{We present Large Motion Model (\name), the first generalist multi-modal motion generation model, that can perform multiple motion generation tasks simultaneously and achieve competitive performance across nine widely used benchmarks.}
    \label{fig:teaser}
    \vspace{-30pt}
\end{figure}

%% file: secs/0_abstract.tex
\begin{abstract}

Human motion generation, a cornerstone technique in animation and video production, has widespread applications in various tasks like text-to-motion and music-to-dance. Previous works focus on developing specialist models tailored for each task without scalability. In this work, we present \textbf{Large Motion Model (LMM)}, a motion-centric, multi-modal framework that unifies mainstream motion generation tasks 
into a generalist model. A unified motion model is appealing since it can leverage a wide range of motion data to achieve broad generalization beyond a single task. However, it is also challenging due to the heterogeneous nature of substantially different motion data and tasks. LMM tackles these challenges from three principled aspects: \textbf{1)} \textit{Data:} We consolidate datasets with different modalities, formats and tasks into a comprehensive yet unified motion generation dataset, \textbf{MotionVerse}, comprising 10 tasks, 16 datasets, a total of 320k sequences, and 100 million frames. \textbf{2)} \textit{Architecture:} We design an articulated attention mechanism \textbf{\attname} that incorporates body part-aware modeling into Diffusion Transformer backbone. \textbf{3)} \textit{Pre-Training:} We propose a novel pre-training strategy for LMM, which employs variable frame rates and masking forms, to better exploit knowledge from diverse training data. Extensive experiments demonstrate that our generalist LMM achieves competitive performance across various standard motion generation tasks over state-of-the-art specialist models. Notably, LMM exhibits strong generalization capabilities and emerging properties across many unseen tasks. Additionally, our ablation studies reveal valuable insights about training and scaling up large motion models for future research.
  
  \keywords{Motion Generation \and Unified Model \and Multi-Modality}
\end{abstract}

%% file: secs/1_intro.tex
\section{Introduction}
\label{sec:intro}

Humans perform a variety of motions in response to environmental changes, personal thoughts, and emotions to achieve their goals. These intricate motions often serve as the most information-rich element within videos and animations that feature characters, making motion generation a critical component of Generative AI, significantly influencing visual experiences and content quality. Automated human motion generation, aimed at producing continuous, natural, and logical human movements based on specific commands and control conditions, has attracted considerable attention in the computer vision field. 

These specific sub-tasks are characterized by their defined inputs and outputs, with clear objectives. For instance, action-to-motion generates movements based on the category of the motion~\cite{guo2020action2motion,petrovich2021action}; text-to-motion takes textual descriptions and produces corresponding movements~\cite{guo2022generating,zhang2024motiondiffuse}; music-to-dance creates dance moves in tune with the style and beat of the music input~\cite{siyao2022bailando,gong2023tm2d}. Dividing tasks allows for a more focused approach to each sub-task: constructing dedicated datasets and devising methods tailored to the task at hand. However, these approaches, when designed for a singular sub-task or modality, face challenges due to limited data quantity and a narrow data domain, which in turn can lead to models with restricted capabilities and poor generalization performance. In contrast, the objective of this work is to build a unified yet versatile foundation model for human motion generation, leveraging resources from a wide range of applications and achieving strong performance across the board.

Leveraging multi-modal and multi-task motion generation datasets presents significant challenges. First, disparate datasets feature varying motion formats and evaluation metrics, such as keypoint-based or rotation-based formats, and metrics assessing realism or diversity. Consequently, employing a single model to tackle multiple tasks and perform evaluations across different datasets is exceedingly difficult. Furthermore, transferring motion knowledge across tasks within these datasets is challenging, complicating the model's ability to integrate useful knowledge from various data sources to enhance its capabilities. For instance, differences in frame rates and the number of keypoints (sometimes even missing parts of the body) make it hard to unify the learned knowledge. Although some studies attempt to address multiple tasks simultaneously, they often utilize only two or three datasets with the same motion format, which limits their ability to achieve enhanced controllability and generalizability. In summary, integrated motion generation models for multi-modal and multi-task applications encounter the following problems: 1) Non-uniformity of motion data formats; 2) Different evaluation metrics due to varying task objectives; 3) Difficulty in transferring action knowledge across multiple tasks.

To deal with these challenges, we first amass multiple cross-modal motion datasets, encompassing 16 datasets with a total of 320k sequences and 100 million frames. These datasets span seven standard tasks: text-to-motion, action-to-motion, motion prediction, speech-to-gesture, music-to-dance, motion imitation, and motion in-betweening. Additionally, based on the standard tasks and multi-modal control signals, we introduce three new tasks: conditional motion prediction, conditional motion in-betweening, and multi-condition motion generation. Together, these datasets and tasks form our cross-modal motion benchmark, \textbf{MotionVerse}. To align the diverse formats of motion data, we employ a two-step approach: 1) We use the TOMATO representation~\cite{lu2023humantomato} as a unified intermediary format, and then divide the entire representation into 10 parts. All kinds of motion representations are aligned to this format, with annotations indicating which body parts are present in each sequence; 2) We train a series of representation translators to convert the unified motion representation into the specific representations of each dataset during the testing phase. With MotionVerse, we can smoothly use training data from various tasks and modalities, and conduct tests across different datasets.

Building on MotionVerse, we introduce the multi-modal Large Motion Model (\textbf{LMM}), which is built on a transformer-based diffusion model. Addressing the motion format inconsistency,
we developed a body part-aware motion generation model. This model divides the human body into 10 segments and employs a specialized attention mechanism \textbf{\attname}, featuring multi-conditioning, spatial-temporal independence, and mask injection,
allowing for distinct control over different body parts. Furthermore, body part-aware modeling decomposes motion data from various datasets into relatively independent segments, thereby enabling the model to more effectively leverage knowledge learned across different datasets. Lastly, we adopt learning strategies from large language models (LLM), proposing a training method for LMM that combines unsupervised and supervised learning. In unsupervised learning, we enhance model robustness to frame rates through random frame rate augmentation and improve control over the continuity of body part movements by applying random masks to sequences and body parts in various ways. This training approach significantly leverages large amounts of multi-modal data to bolster LMM's capabilities. In supervised learning, we refine the capabilities of models to enhance their performance on specific tasks.
%
%
Experimental results show that LMM achieves state-of-the-art results across various tasks, demonstrating its exceptional generalization performance, as shown in Fig.~\ref{fig:teaser}. Furthermore, LMM has the ability to process multi-modal inputs simultaneously, enabling it to accomplish unseen tasks.
%

In summary, our core contributions are as follows:

\textbf{1.} We present MotionVerse, a mega-scale, multi-modal, multi-task motion generation dataset that features a unified motion representation across a wide range of tasks and motion formats.

\textbf{2.} We introduce a Large Motion Model (LMM) that incorporates an advanced attention mechanism \attname, allowing for precise and robust control, achieving finer results.


\textbf{3.} We devise a pre-training strategy for the LMM, including random frame rates and various masking techniques to fully leverage extensive motion datasets and enhance the model's capabilities. 
Additionally, through ablation studies on our training approach, we explored certain characteristics inherent in LMM's training process, laying a foundation for future research on the LMM.

%% file: secs/2_related.tex
\section{Related Work}
\label{sec:related}

\subsection{Motion Synthesis}

Subtasks of motion generation are differentiated based on the types of control signals they utilize. Some tasks require the algorithm to synthesize motion sequences based on the given uncompleted motion sequences. For example, in the motion prediction task~\cite{gopalakrishnan2019neural, sun2021action, guo2023back, barquero2023belfusion, ahn2023can, DiffPred:2023, diller2022charposes, yan2023gazemodiff, wang2023gcnext, mao2020history, barsoum2018hp, chen2023humanmac, jiang2023motiondiffuser, sun2023towards}, the control condition is the first several poses, and the generated motion must logically follow the preceding sequence to ensure the extended sequence appears natural and reasonable. There are also works focusing on recovering the whole body motion with the given sparse upper-body tracking signals~\cite{du2023avatars, castillo2023bodiffusion}.

Action-to-motion~\cite{guo2020action2motion, petrovich2021action, cervantes2022implicit, kalakonda2023action, liu2023language, wang2020learning, zhao2023modiff, kulal2022programmatic} is an inverse task derived from the motion recognition task. While the latter identifies the action category from a motion sequence, the former involves generating a corresponding action sequence from an input action category. 
In the music-to-dance task~\cite{huang2020dance, li2020learning, li2021ai,  zhuang2022music2dance, siyao2022bailando, okamura2023dance, yao2023dance, gao2023dancemeld, qi2023diffdance, tseng2023edge, yang2023longdancediff, lee2019dancing, sun2020deepdance, li2022danceformer}, the control signal is music, and the motion sequence (dance) should be in accordance with the style and rhythm of the music. 
Bailando~\cite{siyao2022bailando} addresses crucial spatial constraints and temporal coherence in dance generation by utilizing a codebook to store standardized dancing units, confining spatial constraints. It achieves temporal fluidity through a GPT designed to detect music beats.
Another task based on audio is speech-to-gesture~\cite{kucherenko2019analyzing, xu2023chain, yin2023emog, Ao2023GestureDiffuCLIP, liu2022learning, zhi2023livelyspeaker, yoon2020speech, ghorbani2023zeroeggs}, where the input is the speaker's audio and the output is the corresponding gestures of the speaker, taking into account the speech's pauses, emotional fluctuations, etc. 
GestureDiffuCLIP~\cite{Ao2023GestureDiffuCLIP} utilizes CLIP to extract style information from the input and then employs a diffusion model to generate gestures.
Text-to-motion is one of the most attracting topic in conditional motion generation~\cite{lin2018generating, ahuja2019language2pose, ghosh2021synthesis, petrovich2022temos, tevet2022motionclip, tevet2022human, kim2023flame, athanasiou2022teach, jin2023act, jing2023amd, han2023amd, zhong2023attt2m, lin2023being, qian2023breaking, lou2023diversemotion, zhou2023emdm, wei2023enhanced, chen2023executing, wang2023fg, zhang2024finemogen, DBLP:conf/starsem/ZhangTSKGM19, shi2023generating, karunratanakul2023guided, shafir2023human, lu2023humantomato, ren2024insactor, goel2023iterative, zhai2023language, DBLP:conf/iccv/AzadiSHPG23, ling2023mcm, pinyoanuntapong2023mmm, yao2023moconvq, dabral2023mofusion, guo2023momask, hu2023motion, zhang2024motiondiffuse, zhang2023motiongpt, jiang2024motiongpt, ribeiro2024motiongpt, hoang2024motionmix, yazdian2023motionscript, petrovich2024multi, lin2022ohmg, xie2023omnicontrol, yuan2023physdiff, liu2023plan, kong2023priority, zhang2023remodiffuse, he2023semanticboost, li2023sequential, DBLP:conf/iccv/AthanasiouPBV23, raab2023single, liu2023spatio, DBLP:conf/siggrapha/QingCYY23, yang2023synthesizing, zhang2023generating, zhang2023tapmo, ghosh2021text, wan2023tlcontrol, gong2023tm2d, guo2022tm2t, xie2023towards, zhou2023ude, voas2023best}, where motions are generated based on textual descriptions. This requires the model to comprehend the meaning of the text and produce a corresponding sequence of motions. Previous works attempted to apply advanced generative model~\cite{tevet2022human,zhang2024motiondiffuse,zhang2023motiongpt} to improve performance. While some other works focused on enhancing controllability~\cite{qian2023breaking,zhang2024finemogen,DBLP:conf/iccv/AthanasiouPBV23}. Physical reality~\cite{yuan2023physdiff,ren2024insactor} and out-of-domain performance~\cite{tevet2022motionclip,hong2022avatarclip,liu2023plan} are also vital topics in this field. 

In addition, some works focus on human-scene interaction generation~\cite{huang2023diffusion, lim2023mammos, liu2023revisit, xiao2023unified}, human-object interaction generation~\cite{diller2023cghoi, li2023controllable, zhang2022couch, tendulkar2023flex, hao2024hand, lin2023handdiffuse, pi2023hierarchical, peng2023hoi, shimada2023macs, kulkarni2023nifty, li2023object, wang2023physhoi} and human-human interaction generation~\cite{chopin2024bipartite, zhao2023diffugesture, cai2023digital, siyao2023duolando, liu2023interactive, liang2023intergen, ghosh2023remos, tanaka2023role}. These tasks greatly expanded the scope of motion generation applications.

However, action generation targeting a single task often struggles with limited data volume and a singular data domain, leading to models with restricted capabilities and poor generalization. Our paper integrates various motion generation tasks, designing the Large Motion Model (LMM) to utilize multi-modality data from different tasks for model training. This enables the model to learn from various domains, enhancing its generalization capabilities. 

\subsection{Large Diffusion Model}


As diffusion models have made remarkable strides in image generation tasks, researchers have extended their application to a broader array of fields, including video generation~\cite{kim2023diffusion, voleti2022mcvd, yang2023rerender}, image editing~\cite{hertz2022prompt, brooks2023instructpix2pix, couairon2022diffedit, kawar2023imagic, nguyen2023visual}, and motion generation~\cite{dabral2023mofusion, zhang2024motiondiffuse,zhang2024finemogen,zhou2023emdm,tevet2022human}, etc. Moreover, given the limited functionality and control over generation provided by single-modal conditions, multi-modal inputs have been introduced into diffusion models to enhance their versatility and control capabilities. In the realm of image generation, UNIMO-G~\cite{li2024unimo} takes both images and text as inputs, utilizing the subjects in pictures and textual prompts to generate realistic images that match complex semantics.
For video generation, MM-Diffusion~\cite{ruan2023mm} incorporates audio and video in a multi-modal manner, enabling the harmonious adaptation of audio and visuals to produce realistic videos with sound.
In image editing tasks, the integration of multi-modal inputs with diffusion models has made image editing more flexible and convenient. 
Control-color~\cite{liang2024control} combines text, strokes, exemplars, and other conditions to achieve interactive, multi-modal controlled image coloring. 
InstructAny2pix~\cite{li2023instructany2pix} uses similar conditions for multi-modal control over inpainting.
Likewise, in the field of motion generation, works like MotionDiffuse~\cite{zhang2024motiondiffuse} and MDM~\cite{tevet2022human} have tackled text-to-motion and action-to-motion. MCM~\cite{ling2023mcm}, UDE~\cite{zhou2023ude} has addressed text-to-motion and music-to-dance. In summary, across different domains, multi-modal diffusion models enrich the content and enable more precise control of generation tasks. Previous dual-modal motion generation methods often utilize similar annotation formats, such as SMPL~\cite{loper2015smpl}, facilitating easy data alignment. However, the multi-modal datasets we collected cover a broader domain span, and the formats for motion annotation are extremely diverse. Thus we propose a comprehensive benchmark, \dname, to unify the motion representation, ultimately leading to the development of LMM.


%% file: secs/3_method.tex
\section{MotionVerse: Unified Motion Generation Datasets}

\subsection{Motivation}

Large models have been extensively studied in fields such as language, image, and video. These models, by absorbing common knowledge from vast amounts of data and leveraging a unified task format, demonstrate outstanding performance across multiple tasks. However, on the path towards large motion models, a significant challenge lies ahead: the inconsistent motion formats across datasets. Specifically, there are three types of inconsistencies:
\begin{enumerate}
    \item \textbf{Inconsistent pose representations}: For instance, the UESTC~\cite{ji2018large} benchmark adopts a 6D rotation representation based on SMPL~\cite{SMPL:2015}, while the Human3.6M~\cite{ionescu2013human3} motion prediction benchmark uses keypoint coordinates.
    \item \textbf{Inconsistent number of keypoints}: For example, TED-Gesture++~\cite{yoon2020speech} only includes upper body keypoints, while NTU-RGBD 120~\cite{shahroudy2016ntu,liu2019ntu} lacks fine-grained keypoints for the hands.
    \item \textbf{Inconsistent frame rates}: KIT-ML~\cite{plappert2016kit} operates at 12.5 fps, whereas Motion-X~\cite{lin2023motionx} runs at 30 fps.
\end{enumerate}
Such differences not only demand models capable of handling diverse data formats but also pose significant challenges in acquiring common knowledge.

To address this challenge, we introduce the first unified and comprehensive motion-centric benchmark \textbf{\dname}, the workflow of which is shown in Fig.~\ref{fig:motionverse}. \dname possesses three advantages: 
\begin{enumerate}
    \item \textbf{Unified Problem Formulation}: we describe mainstream tasks within a unified framework, reducing the need to consider task-specific properties during model design.
    \item \textbf{Unified Motion Representation}: we convert the motion formats of various datasets into a unified intermediate representation, enabling the model to acquire common knowledge from the originally diverse data formats and to be evaluated on different datasets smoothly.
    \item \textbf{Systematicness and Comprehensiveness}: we encompass 10 tasks across 16 datasets, comprising 320K sequences and nearly 100M frames of motion data, as shown in Tab.~\ref{tab:dataset_info}, which enables us to explore large motion models.
\end{enumerate}


\input{tables/task}
\subsection{Unified Problem Formulation}
\label{subsec:probdef}

Motion generation encompasses a variety of sub-tasks with differing objectives. To standardize the data for these tasks, we first formalize the input format for motion generation tasks as: 
\begin{equation}\label{eq:task-form}
    \Theta=M(\mathbf{x}, \text{m}, \text{c}), 
\end{equation}
where $\mathbf{x} \in \mathbb{R}^{F \times D}$ represents the motion data, $\text{m} \in \{0,1\}^{F \times D}$ defines the model's visibility scope, which is used in motion completion tasks, such as motion prediction and motion in-betweening. Here $F$ is the number of frames of the target motion sequence. $D$ is the dimensionality of each pose state. $\text{c}$ is a set of condition control signals, including text, speech, music, and video.

We have included seven standard tasks: action-to-motion (A2M), text-to-motion (T2M), music-to-dance (M2D), speech-to-gesture (S2G), motion prediction (MP), motion in-betweening (MIn) and motion imitation (MIm), along with three multi-modal tasks: conditional motion prediction (CMP), conditional motion in-betweening (CMI), and multi-condition motion generation (MMG). For each specific task, we can align their inputs using Eq.\eqref{eq:task-form}. Tab.~\ref{tab:task_def} lists the details of these ten tasks.
    
\input{figures/motionverse}

\subsection{Unified Motion Representation}
\label{subsec:data}


For data formats with varying inputs and outputs across different tasks, we preprocess to ensure format consistency. The basic unit of general motion datasets can be defined as <input-output> pairs. For inputs, we consider multi-modalities including text, speech, music, and video. To align these modalities, we employ Imagebind~\cite{Girdhar_2023_CVPR} to encode different inputs into unified features of the same dimension, which ensures semantic consistency across modalities. 

As for the output, it encompasses motion sequences in various formats, such as keypoints and SMPL~\cite{SMPL:2015}. To standardize motion representation, we define a unified format similar to TOMATO~\cite{lu2023humantomato}. Our representation is described as:
\begin{equation}
    m_{i}=\left\{\dot{r}^{a}, \dot{r}^{x}, \dot{r}^{z}, r^{y}, \mathbf{j}^{p}, \mathbf{j}^{v}, \mathbf{j}^{r}, \mathbf{f}\right\},
\end{equation}
where $r$ denotes information related to the root. Specifically, $\dot{r}^{a} \in \mathbb{R}$ is the angular velocity along the Y-axis, $(\dot{r}^{x},\dot{r}^{z} \in \mathbb{R})$ represent linear velocities on the XZ-plane, and $r^{y}$ indicates the root's height. $\mathbf{j}^{p} \in \mathbb{R}^{3(J-1)}$, $\mathbf{j}^{v} \in \mathbb{R}^{3J}$, and $\mathbf{j}^{r} \in \mathbb{R}^{6(J-1)}$ correspond to the position, velocity, and rotation of local keypoints relative to the root, with $J$ denoting the number of joints ($J-1$ means all joints without the root joint). Here we follow SMPL-X~\cite{SMPL-X:2019} and consider 22 main body joints and 30 hand joints. Lastly, $\mathbf{f}$ denotes facial expression~\cite{li2017learning}. 

We further divide this representation into ten independent parts: global orientation and trajectory, face expression, head, spine, left arm, right arm, left leg, right leg, left hand, and right hand. When processing raw data, we allow for missing body parts and annotate them in the metadata. For missing keypoints, we utilize prior knowledge of the human body for completion, while extra keypoints are discarded in this process. We then train a series of motion translators to map our unified motion representation to each dataset's specific one. Thus, in the testing process of different tasks, once the model outputs in our unified format, we can map the output to the corresponding motion format through the translator, facilitating smooth metric evaluation.

\input{figures/pipeline}

\section{Large Motion Model}
\label{subsec:arch}

Our model architecture closely follows the literature~\cite{tevet2022human,zhang2024motiondiffuse}, built upon a transformer-based diffusion model. We primarily reference the FineMoGen~\cite{zhang2024finemogen} as our baseline and extend it to support various condition signals, multi-tasking, multiple frame rates, and various mask forms. The overall workflow is illustrated in Fig. ~\ref{fig:pipeline} while the detailed architecture is shown in Fig.~\ref{fig:arch}.

\subsection{Transformer-based Diffusion Model}

The diffusion model is a powerful generative model, capable of producing high-quality, diverse outcomes, garnering widespread attention, and demonstrating formidable generative capabilities in many fields. 
Its essence lies in two intertwined processes: the forward diffusion process and the reverse diffusion process. More details are provided in the supplementary materials.


\input{figures/arch}

\subsection{Read-In Layer and Read-Out Layer}
The read-in layer is responsible for transforming noised motion data into feature representations of the form $F \times H \times D$, while the read-out layer generates clean motion data from the feature space. Here, the first dimension represents the number of frames in the motion sequence, and the second dimension represents the number of body parts. Although we standardize all motion data into a unified motion format, the differences in distribution among various datasets cannot be completely ignored. Therefore, in the backbone network's read-in and read-out stages, we employ dataset-dependent motion encoders and decoders. Additionally, to obtain more comprehensive knowledge for practical applications, during training, there is a 10\% probability of replacing the original dataset name with ``all''. Consequently, the corresponding read-in and read-out layers can be better applied in real-world application scenarios.

\subsection{ArtAttention}
To achieve outstanding zero-shot continuous generation capability, our model builds upon the SAMI module from FineMoGen~\cite{zhang2024finemogen}, incorporating upgrades to address three new requirements: multi-modal condition, various frame rates, and allowance for missing body parts, as shown in Fig.~\ref{fig:arch}.

For multi-modal signals, we preprocess all signals into token sequences using the ImageBind~\cite{Girdhar_2023_CVPR} model. To better integrate these features into our network, we further refine them with two learnable transformer encoder layers. This process transforms text, speech, music, and video into feature sequences $\mathbf{C}_t \in \mathbb{R}^{L_t \times (H \cdot D)}, \mathbf{C}_s \in \mathbb{R}^{L_s \times (H \cdot D)}, \mathbf{C}_m \in \mathbb{R}^{L_m \times (H \cdot D)}, \mathbf{C}_v \in \mathbb{R}^{L_v \times (H \cdot D)}$, where $L_t,L_s,L_m,L_v$ represent the lengths of the corresponding condition sequences, and $H \cdot D$ denotes the feature length of each element.

Our attention mechanism can be divided into two main components: body-part attention(spatial attention) and temporal attention. Assuming the feature representation of our motion sequence is $\mathbf{X} \in \mathbb{R}^{F \times H \times D}$. In the body-part attention section, due to the presence of inherent missing body parts in our data and the masked body parts introduced artificially during pre-training, we cannot utilize a fixed set of coefficients to determine the mutual contributions among body parts. Therefore, unlike the design in FineMoGen, for each frame, we opt to use an attention structure to obtain a set of refined features $\mathbf{Y_s} \in \mathbb{R}^{F \times H \times D}$.

In the temporal attention section, we aim to leverage the self-correlation inherent in the motion features $\mathbf{X}$ and guidance obtained from the condition signals $\mathbf{C}_t,\mathbf{C}_s,\mathbf{C}_m,\mathbf{C}_v$ to generate higher-quality features for each body part in every frame. Here, we employ a Multi-head Attention mechanism, with each head focusing on a specific body part, emphasizing the utilization of temporal correlations to optimize features. We begin by utilizing Mixture-of-Expert to obtain a set of features $\mathbf{K} \in \mathbb{R}^{L \times D}$ from these condition features, where $L$ represents the sequence length of the corresponding feature source. 

Empirically, we found that directly concatenating motion sequences and condition sequences like FineMoGen and then applying Softmax processing hinders the modeling of self-correlation in multi-condition scenarios, resulting in lower motion quality, especially when the condition feature sequence is much longer than the motion sequence. Therefore, we independently normalize the motion feature $\mathbf{K}_x \in \mathbb{R}^{F \times D}$ obtained and then normalize condition features. To support unconditional generation, we introduce 64 learnable tokens as placeholders. These tokens are concatenated with all condition signals for normalization, resulting in $\mathbf{K}_c \in \mathbb{R}^{(64+L_t+L_s+L_m+L_v) \times D}$. This approach also facilitates better blending of the model across different conditions. The remaining processing is similar to FineMoGen. After obtaining a series of time-varying signals, for each frame's pose, we use time as the sole query feature to calculate the correlation between each frame and each time-varying signal, as well as the values at each signal point. Unlike FineMoGen, which uses frame indices to represent time, we use real time to support different frame rates. More details are introduced in the supplementary material. Suppose the output of the temporal attention section is $\mathbf{Y}_t \in \mathbb{R}^{F \times H \times D}$, then the output of the entire \attname is $\mathbf{Y}=\mathbf{Y}_s+\mathbf{Y}_t$.


\subsection{Pre-Training and Fine-Tuning}
\label{subsec:train}

Leveraging \dname, we have collected a vast array of <input-motion> data pairs. Disregarding the inputs, the abundant motion data inherently contains valuable information, which can enable the model to comprehend numerous characteristics of human motion, such as coherence, balance, and joint-based rotations, among others. To enable the model to better acquire common knowledge across datasets, we divide the entire training process into two main parts: \textbf{Unsupervised Pre-Training} and \textbf{Supervised Fine-Tuning}. In the first stage, we require our model to learn motion priors independent of conditions. While in the second stage, the model is supposed to learn the correlation between condition signals and motion sequences.

As illustrated in Fig.~\ref{fig:pipeline}, in the unsupervised pretraining phase, to enrich the model's prior knowledge from these motion sequences, we employ random downsampling and random masking strategies for data augmentation. Since our representation includes terms related to velocity, when downsampling, we need to recalculate the velocity values to match the downsampling rate, while the terms related to states remain unchanged. This approach enables the model to better learn from data with different original frame rates. As mentioned earlier in the data preprocessing part, some body parts in the sequences are masked out, such as the detailed keypoints of the left and right hands in the KIT dataset. We denote this original mask as $\mathbf{M}_s \in \{0,1\}^{F \times H}$. Based on this, we additionally apply masking with a certain probability to obtain a new mask $\mathbf{M}_t \in \{0,1\}^{F \times H}$. After the model performs the read-in operation, we replace the body parts marked as 
$1$ in $\mathbf{M}_t$ with learnable empty tokens. When calculating the loss, we only ignore the parts marked by $\mathbf{M}_s$. This means that the model not only needs to restore the noised sequence to its clean parts but also utilize the visible part to infer the rest. With this modeling, the knowledge embedded in the data with missing body parts can be better absorbed by the model.

In the supervised fine-tuning phase, our goal is to enable the model to learn the relationship between condition signals and motion sequences. Here, we pass the preprocessed condition token sequences as additional inputs to the model. To support classifier-free guidance, during training, we randomly mask out the condition signals with a probability of 10\%.

%% file: tables/task.tex
\begin{minipage}{\textwidth}
    \begin{minipage}{0.42\textwidth}
        \centering
        \resizebox{1\linewidth}{!}{\begin{tabular}{l|c|c}
        \hline
        \textbf{Task}  & \textbf{Mask} & \textbf{Condition}  \\
        \hline
        \hline
        T2M  & None & Text \\
        A2M  & None & Action \\
        M2D  & None & Music \\
        S2G  & None & Speech \\
        MIm  & None & Video  \\
        MP  & $m=
        \begin{cases}
         0 & \text{ if } x>k \\
         1 & \text{ Otherwise}
        \end{cases}$ & None \\
        MIn  & $m=
        \begin{cases}
         0 & \text{ if } k_{1}<x \le k_{2} \\
         1 & \text{ Otherwise}
        \end{cases}$ & None  \\
        \hline
        CMP  & $m=
        \begin{cases}
         0 & \text{ if } x>k \\
         1 & \text{ Otherwise}
        \end{cases}$ & Any single modal \\
        CMI  & $m=
        \begin{cases}
         0 & \text{ if } k_{1}<x \le k_{2} \\
         1 & \text{ Otherwise}
        \end{cases}$ & Any single modal  \\
        MMG  & None & Multi-modal  \\
        \hline
        \end{tabular}}
        \captionof{table}{\textbf{Task definitions.} $x$ and $k_*$ are the frame indices, and $k_*$ represents the boundary of the mask.}
        \label{tab:task_def}
    \end{minipage}
    \begin{minipage}{0.02\textwidth}
        
    \end{minipage}
    \begin{minipage}{0.56\textwidth}
        \centering
        \resizebox{0.9\linewidth}{!}{\begin{tabular}{l|c|c|c|c}
        \hline
        \textbf{Dataset} & \textbf{\#Seq} & \textbf{\#Frames} & \textbf{Repr} & \textbf{Condition} \\
        \hline
        \hline
        HumanML3D~\cite{guo2022generating}  & 14614 & 2M & H3D & Text \\
        KIT-ML~\cite{plappert2016kit}  & 2485 & 245K & H3D & Text \\
        Motion-X~\cite{lin2023motionx}  & 50863 & 9M & SMPLX & Text \\
        BABEL~\cite{punnakkal2021babel}  & 5123 & 7M & SMPLX & Text \\
        UESTC~\cite{ji2018large}  & 25600 & 10M & SMPL & Action \\
        HumanAct12~\cite{guo2020action2motion}  & 1191 & 90K & Kpt3D & Action \\
        NTU-RGB-D 120~\cite{shahroudy2016ntu,liu2019ntu}  & 139656 & 10M & Kpt3D & Action \\
        AMASS~\cite{AMASS:ICCV:2019} & 14244 & 20M & SMPLX & - \\
        3DPW~\cite{vonMarcard2018} & 81 & 140K & SMPL & Video  \\
        Human3.6M~\cite{ionescu2013human3} & 210 & 530K & Kpt3D & Video  \\
        TED-Gesture++~\cite{yoon2020speech} & 34491 & 10M & Kpt3D & Speech  \\
        TED-Expressive~\cite{liu2022learning} & 27221 & 8M & Kpt3D & Speech  \\
        Speech2Gesture-3D~\cite{kucherenko2021moving} & 1047 & 1M & Kpt3D & Speech  \\
        BEAT~\cite{liu2022beat} & 1639 & 18M & Kpt3D & Speech  \\
        AIST++~\cite{li2021ai} & 1408 & 1M & SMPL & Music \\
        MPI-INF-3DHP~\cite{mono-3dhp2017} & 16 & 1M & Kpt3D & Video  \\
        \hline
        Total & 320K & 100M & - & - \\
        \hline
        \end{tabular}}
        \captionof{table}{\textbf{Dataset information.} We collect 16 widely used dataset, process all motion data into our intermediate format.}
        \label{tab:dataset_info}
    \end{minipage}
\end{minipage}

%% file: figures/motionverse.tex
\begin{figure}[t]
    \centering
    \includegraphics[width=\linewidth]{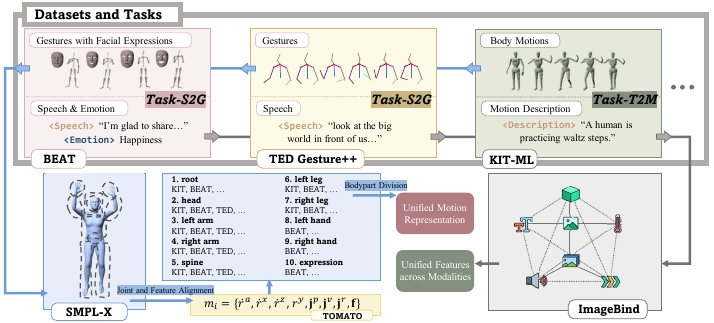}
    \caption{\textbf{\dname}. We preprocess distinct motion-centric datasets into a unified format. As for motion sequences, we initially convert them to the TOMATO~\cite{lu2023humantomato} representation and then further divide them into 10 independent body parts, serving as our unified motion representation. To tackle multi-modal condition signals, we employ ImageBind~\cite{Girdhar_2023_CVPR} to transform them into unified features across modalities.}
    \label{fig:motionverse}
    \vspace{-15pt}
\end{figure}

%% file: figures/pipeline.tex
\begin{figure}[t]
    \centering
    \includegraphics[width=\linewidth]{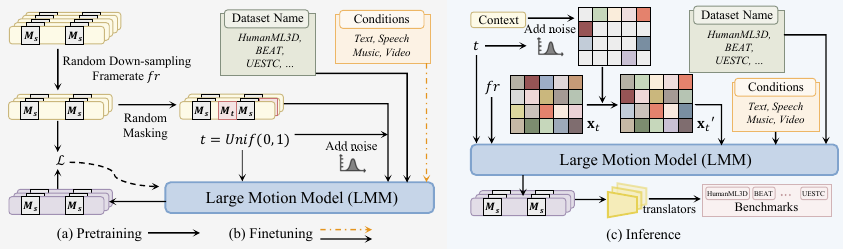}
    \vspace{-10pt}
    \caption{\textbf{Overall pipeline of \name.} \textbf{Left}: Our two-stage training procedure, including unsupervised pretraining and supervised fine-tuning. Random down-sampling and random mask strategies are applied to enhance knowledge absorption. \textbf{Right}: The generic inference process of \name. The noised motion sequence and the given context are initially merged before being input into the network. \name will then synthesize motion sequences, consistent with the provided multi-modal condition signals.}
    \vspace{-15pt}
    \label{fig:pipeline}
\end{figure}

%% file: figures/arch.tex
\begin{figure}[t]
    \centering
    \includegraphics[width=\linewidth]{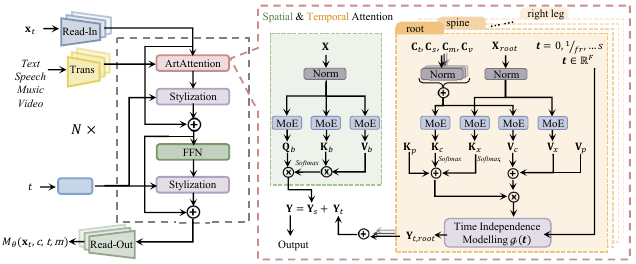}
    \caption{\textbf{Architecture of \name.} \name is a transformer-based diffusion model. Dataset-dependent Read-In layers and Read-Out layers facilitate the conversion of the motion sequence between our intermediate representation and the latent feature space. In the stem of \name, \attname refines the feature representations through the spatial and temporal attention branches.}
    \label{fig:arch}
    \vspace{-15pt}
\end{figure}

%% file: secs/4_experiment.tex
\section{Experiments}



\subsection{Implementation Details} 

We designed four variants: LMM-Tiny, LMM-Small, LMM-Base, and LMM-Large, which have 90M, 160M, 410M, 760M parameters respectively. We use all data in \dname for both pretraing and finetuning, except for the sequences used in evaluation.
We maintained a fixed total batch size of 512. For the Tiny model, we conducted training directly on 8 NVIDIA V100 GPUs with 32GB memory each, with a batch size of 64 per GPU. For larger models, we FP16 and gradient accumulation to achieve training effects equivalent to a batch size of 512 without exceeding 32 V100 GPUs. During the pre-training phase, we employed the Adam optimizer with a fixed learning rate of $2\times10^{-4}$ for 80K iterations. In the fine-tuning phase, we used the same optimizer, initially iterating for 20K steps with a learning rate of $2\times10^{-4}$, followed by 20K steps with a learning rate of $2\times10^{-5}$. For more details, please refer to the supplementary material.

\subsection{Quantitative Results}

\input{tables/human_ml3d}

\input{tables/motion_prediction}

\input{tables/music_dance}

We evaluate our LMM variants on three tasks: text-to-motion, music-to-dance, motion prediction, and four datasets: HumanML3D~\cite{guo2022generating}, 3DPW~\cite{vonMarcard2018}, AMASS~\cite{AMASS:ICCV:2019}, and AIST++~\cite{li2021ai}, as shown in Tab.~\ref{tab:humanml3d}, Tab.~\ref{tab:motion_prediction} and Tab.~\ref{tab:musicdance}. More experimental results are reported in the supplementary material. 

\indent\textbf{Text-to-Motion.} Tab.~\ref{tab:humanml3d} demonstrates that our LMM-Large surpasses other existing works in terms of accuracy and fidelity while maintaining comparable diversity. On the other hand, LMM-Tiny, which shares a similar structure with FineMoGen, performs worse than it. This discrepancy can be attributed to the significant challenges posed by the large amounts of diverse data and the trade-offs across different tasks during model training, especially for smaller models like LMM-Tiny. Additionally, it's worth noting that the representation dimension used in HumanML3D is 263, whereas ours is 669, significantly increasing the learning difficulties for LMM-Tiny.

\noindent\textbf{Motion Prediction.} Tab.~\ref{tab:motion_prediction} reports the performance on AMASS and 3DPW test splits. The superior performance of LMM-Large can be attributed to its robust motion prior, which is obtained from the mega-scale data. It is worth noting that due to the errors introduced by the motion translation step, the accuracy of LMM-Large is still lower than other methods in short-distance prediction. However, it exhibits a significant advantage in long-distance prediction. Furthermore, we observed that the advantage of LMM-Large is more pronounced on 3DPW. This is because the 3DPW benchmark demands higher generalization ability from the model. After extensive learning of motion priors, our LMM-Large exhibits a more prominent performance on out-of-distribution tests.

\noindent\textbf{Music-to-Dance.} Our Large model achieves comparable performance  to the current state-of-the-art, as shown in Tab.~\ref{tab:musicdance}. In terms of diversity-related metrics, our approach demonstrates a significant advantage. Our performance in the metrics $FID_k$ and $FID_g$ did not surpass existing methods. One possible reason could be the relatively small proportion of the music2dance dataset in the current dataset composition, leading to the model not fully grasping the correlation between the music condition and motion.

\subsection{Ablation Study}

\input{tables/ablation}

Tab.~\ref{tab:pretrain_ablation} shows the ablation results. We observed that random masking is a necessary component. When the model's expressive power is strong enough, it can directly recover the clean motion sequence from the noised motion sequence. Consequently, during the fine-tuning stage, our condition signal may not play its expected role. Introducing random masking during training will make it more difficult for the model to solely restore the original sequence from the motion sequence, leading it to rely more on the additional information provided by the condition signal. Additionally, we found that both downsampling and random masking strategies are beneficial for improving the multimodality metrics in the text-to-motion task. This implies that the model can better absorb knowledge from different datasets with the help of these two strategies. These strategies also significantly impact the effectiveness of motion prediction. Finally, we compared our proposed ArtAttention with the original SAMI and found that our proposed method is more suitable for the scenario of large motion models.


\input{figures/viz}

\subsection{Qualitative Results}


As shown in Fig.~\ref{fig:viz} (a)-(d), LMM-Large can response to diverse textual descriptions with fine-grained control, which benefits from the large-scale training data and the well-designed architecture. In addition, Fig.~\ref{fig:viz} (e)-(f) provide examples for motion generation under both text description and music rhythms. Our generated motions successfully execute the given commands and follow the music beats simultaneously. 
For more visualization results, please kindly refer to the demo video in our homepage.

\subsection{More Applications}


\begin{figure}[t]
    \centering
    \includegraphics[width=0.7\linewidth]{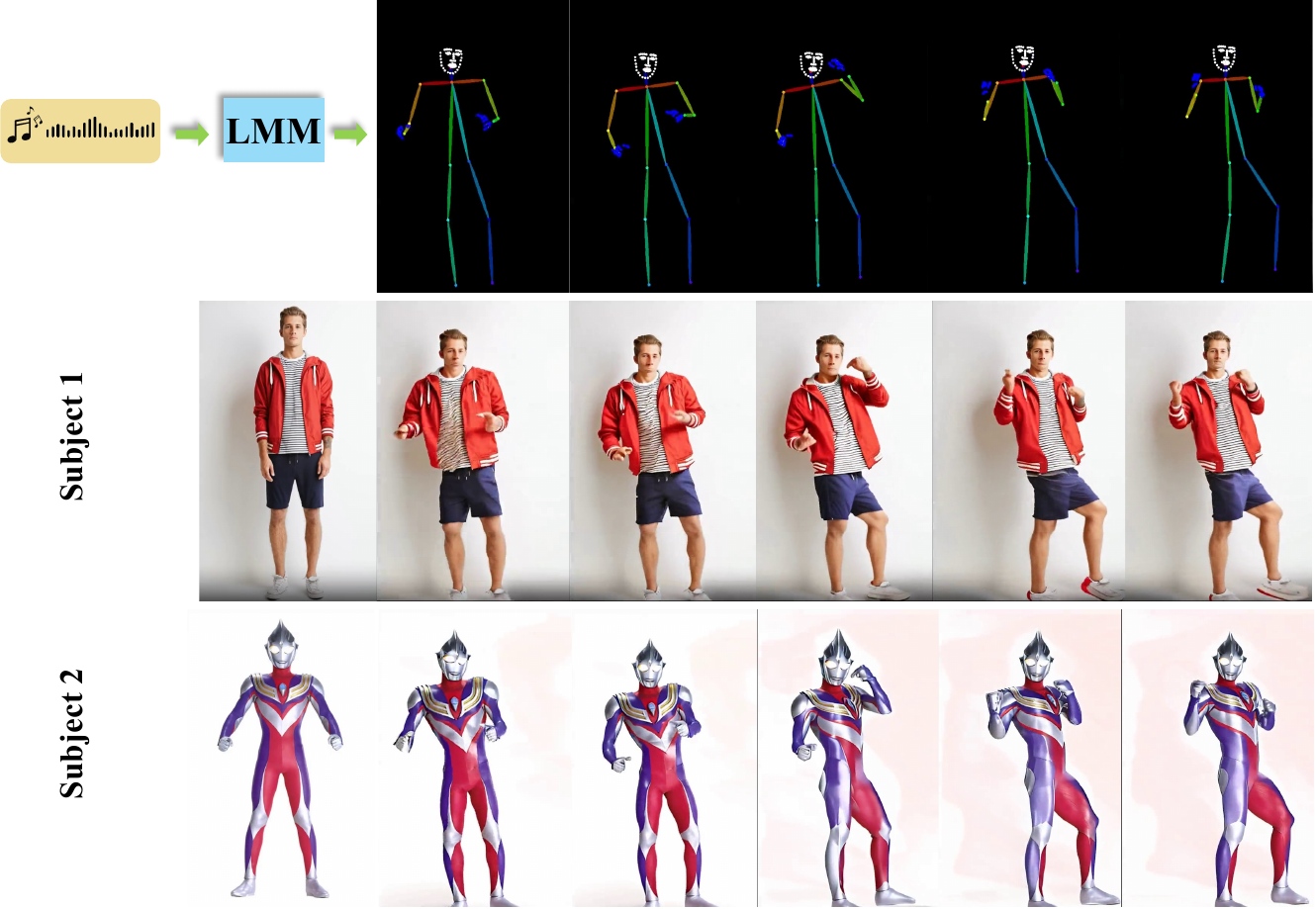}
    \caption{\textbf{Video Generation with our synthesized motion sequence.} After generating a sequence of motions conditioned on music by our LMM-Large, we map the 3D keypoints to a 2D plane, serving as guidance for video generation.}
    \label{fig:animate}
    \vspace{-15pt}
\end{figure}

In Figure ~\ref{fig:animate}, we show two videos that are generated based on our synthesized motion sequences. As a vital application direction, users can leverage our large motion model to customize their desired motion data by providing personalized condition signals, such as text commands or accompanying music. With the assistance of off-the-shelf motion-guided video generation technology, users can freely create videos for their favorite characters.

%% file: tables/human_ml3d.tex
\begin{table}[t]
\tiny
\centering
\caption{\textbf{Quantitative results of text-to-motion generation on the HumanML3D test set.} `$\uparrow$'(`$\downarrow$') indicates that the values are better if the metric is larger (smaller). We run all the evaluations 20 times and report the average metric and 95\% confidence interval. ``MM'' is MultiModality. The best scores are bold, and the second-best results are underlined.}
\vspace{-5pt}

\setlength{\tabcolsep}{0.8mm}
{
\begin{tabular}{lccccccc}
\hline

\multirow{2}{1.3cm}{\centering Methods} & \multicolumn{3}{c}{\centering R Precision$\uparrow$} & \multirow{2}{0.9cm}{\centering FID$\downarrow$} & \multirow{2}{1.1cm}{\centering MM Dist$\downarrow$} & \multirow{2}{0.9cm}{\centering Diversity$\uparrow$} & \multirow{2}{0.9cm}{\centering MM$\uparrow$} \\
& Top 1 & Top 2 & Top 3 \\
\hline
Real motions & $0.511^{\pm .003}$ & $0.703^{\pm.003}$ & $0.797^{\pm.002}$ & $0.002^{\pm.000}$ & $2.974^{\pm.008}$ & $9.503^{\pm.065}$ & -\\ 
\hline






T2M-GPT~\cite{zhang2023generating} & $0.491^{\pm.003}$ & $0.680^{\pm.003}$ & $0.775^{\pm.002}$ & $0.116^{\pm.004}$ & $3.118^{\pm.011}$ & \underline{$9.761^{\pm.081}$} & $1.856^{\pm .011}$ \\

MDM~\cite{tevet2022human} & - & - & $0.611^{\pm.007}$ &  $0.544^{\pm.044}$ & $5.566^{\pm.027}$ & $9.559^{\pm.086}$ & $\mathbf{2.799^{\pm.072}}$ \\


FineMoGen~\cite{zhang2024finemogen} & $0.504^{\pm.002}$ & $0.690^{\pm.002}$ & $0.784^{\pm.002}$ & $0.151^{\pm.008}$ & $2.998^{\pm.008}$ & $9.263^{\pm.094}$ & \underline{$2.696^{\pm.079}$} \\
MoMask~\cite{guo2023momask} & \underline{$0.521^{\pm.002}$} & \underline{$0.713^{\pm.002}$} & \underline{$0.807^{\pm.002}$} & \underline{$0.045^{\pm.002}$} & \underline{$2.958^{\pm.008}$} & - & $1.241^{\pm.040}$ \\
\hline
LMM-Tiny & $0.496^{\pm.002}$ & $0.685^{\pm.002}$ & $0.785^{\pm.002}$ & $0.415^{\pm.002}$ & $3.087^{\pm.012}$ & $9.176^{\pm.074}$ & $1.465^{\pm.048}$ \\
LMM-Small & $0.505^{\pm.002}$ & $0.693^{\pm.002}$ & $0.789^{\pm.002}$ & $0.227^{\pm.002}$ & $3.051^{\pm.012}$ & $9.295^{\pm.076}$ & $1.761^{\pm.049}$  \\
LMM-Base & $0.511^{\pm.002}$ & $0.710^{\pm.002}$ & $0.802^{\pm.002}$ & $0.138^{\pm.002}$ & $2.971^{\pm.012}$ & $9.573^{\pm.076}$ & $2.426^{\pm.054}$  \\
LMM-Large & $\mathbf{0.525^{\pm.002}}$ & $\mathbf{0.719^{\pm.002}}$ & $\mathbf{0.811^{\pm.002}}$ & $\mathbf{0.040^{\pm.002}}$ & $\mathbf{2.943^{\pm.012}}$ & $\mathbf{9.814^{\pm.076}}$ & $2.683^{\pm.054}$  \\
\hline
\end{tabular}}
\vspace{-5pt}
\label{tab:humanml3d}
\end{table}

%% file: tables/motion_prediction.tex
\begin{table}[t]
\tiny
\centering
\caption{\textbf{Quantitative results of motion prediction on the AMASS and 3DPW test set} for different time steps (ms). We report the MPJPE error in \textit{mm}.}
\vspace{-5pt}
\setlength{\tabcolsep}{0.8mm}
{
\begin{tabular}{l|cccccccc|cccccccc}
\hline

\multirow{2}{1.5cm}{\centering Method} & \multicolumn{8}{c|}{\centering AMASS-BMLrub} & \multicolumn{8}{c}{\centering 3DPW} \\
 & 80 & 160 & 320 & 400 & 560 & 720 & 880 & 1000 & 80 & 160 & 320 & 400 & 560 & 720 & 880 & 1000  \\
\hline
LTD-10-10~\cite{mao2019learning} & \underline{10.3} & \textbf{19.3} & 36.6 & 44.6 & 61.5 & 75.9 & 86.2 & 91.2 & \underline{12.0} & \textbf{22.0} & 38.9 & 46.2 & 59.1 & 69.1 & 76.5 & 81.1 \\
SIMLPE~\cite{guo2023back} & 10.8 & \underline{19.6} & \underline{34.3} & 40.5 & \underline{50.5} & \underline{57.3} & 62.4 & 65.7 & 12.1 & \underline{22.1} & 38.1 & 44.5 & 54.9 & 62.4 & 68.2 & 72.2 \\
GCNext~\cite{wang2023gcnext} & \textbf{10.2} & \textbf{19.3} & \textbf{34.1} & \underline{40.3} & 50.6 & \underline{57.3} & \underline{62.0} & \underline{65.3} & \textbf{11.8} & \textbf{22.0} & \underline{37.9} & 44.2 & \underline{55.1} & \underline{62.1} & \underline{67.8} & \underline{72.0} \\
\hline
LMM-Tiny & 15.9 & 24.1 & 38.2 & 45.9 & 61.2 & 73.4 & 80.3 & 87.2 & 17.3 & 26.2 & 40.1 & 47.3 & 62.8 & 74.8 & 82.5 & 87.0 \\
LMM-Small & 14.7 & 23.2 & 37.5 & 43.8 & 58.3 & 69.2 & 75.2 & 81.9 & 16.2 & 25.7 & 39.4 & 45.9 & 60.7 & 71.5 & 79.5 & 82.8 \\
LMM-Base & 13.1 & 21.5 & 35.9 & 41.1 & 53.6 & 60.8 & 66.9 & 70.3 & 14.1 & 23.9 & 38.2 & \underline{44.1} & 55.3 & 64.8 & 70.3 & 73.6 \\
LMM-Large & 12.8 & 20.9 & \underline{34.3} & \textbf{39.6} & \textbf{49.1} & \textbf{55.3} & \textbf{60.5} & \textbf{63.1} & 13.1 & 22.6 & \textbf{37.1} & \textbf{42.4} & \textbf{52.4} & \textbf{59.2} & \textbf{63.8} & \textbf{68.0} \\
\hline
\end{tabular}}
\vspace{-10pt}
\label{tab:motion_prediction}
\end{table}

%% file: tables/music_dance.tex
\begin{table}[t]
\tiny
\centering
\caption{\textbf{Quantitative results for Music-conditioned Dance Generation.} Quantitative results on AIST++ test set.}
\vspace{-5pt}
\label{tab:musicdance}
\setlength{\tabcolsep}{1mm}
{
\begin{tabular}{l|ccccccc}
\hline

\multirow{2}{2.2cm}{\centering Methods} & \multicolumn{2}{c}{\centering Motion Quality} & \multicolumn{2}{c}{\centering Motion Diversity} & \multicolumn{2}{c}{\centering Freezing} & \multirow{3}*{\centering Best Align Score$\uparrow$} \\


 & $\mathrm{FID}_{k}$ $\downarrow$ & $\mathrm{FID}_{g}^{\dag}$ $\downarrow$ & $\mathrm{Div}_{k}$ $\uparrow$ & $\mathrm{FID}_{g}^{\dag}$ $\uparrow$ & PFF$\downarrow$ & $\mathrm{AUC}_{f}$ $\downarrow$ & \\

\hline

Ground-truth & 17.10 & 10.60 & 8.19  & 7.45 & 0.00 & 0.00 & 0.2374 \\ 

\hline

DanceNet~\cite{zhuang2022music2dance} & 69.18 & 25.49 & 2.86 & 2.85  & \textbf{0.00} & \underline{0.98} & 0.1430 \\

DanceRevolution~\cite{huang2020dance} & 73.42 & 25.52 & 3.52  & 4.87 & \underline{11.01} & 12.22 & 0.1950 \\
Bailando~\cite{siyao2022bailando} & 28.16 & \textbf{9.62}  & 7.83  & 6.34 & 14.91 & 13.25 & \textbf{0.2332} \\

TM2D\cite{gong2023tm2d} & \textbf{19.01} & \underline{20.09} & \underline{9.45} & 6.36 & \textbf{0.00} & \textbf{0.00} & 0.2049 \\

\hline
LMM-Tiny & 37.62 & 28.95 & 6.92 & 5.94 & \textbf{0.00} & \textbf{0.00} & 0.1736 \\
LMM-Small & 34.18 & 27.53 & 7.46 & 6.17 & \textbf{0.00} & \textbf{0.00} & 0.1791 \\
LMM-Base & 25.43 & 24.18 & 9.05 & \underline{6.55} & \textbf{0.00} & \textbf{0.00} & 0.2084 \\
LMM-Large & \underline{22.08} & 21.97 & \textbf{9.85} & \textbf{6.72} & \textbf{0.00} & \textbf{0.00} & \underline{0.2249} \\
\hline
\end{tabular}}
\vspace{-0pt}
\end{table}

%% file: tables/ablation.tex
\begin{table}[ht]
\centering
\tiny
\caption{\textbf{Ablation of the pretraining strategy.} All experiments utilized LMM-Base as the base model.}
\vspace{-5pt}
\setlength{\tabcolsep}{1mm}
{
\begin{tabular}{c|ccc|ccc|ccc}
\hline
\multirow{2}{0.2cm}{} & \multirow{2}{1.5cm}{Downsample } & \multirow{2}{1.5cm}{Random Mask} &   \multirow{2}{1.1cm}{Attention} & \multicolumn{3}{c|}{\centering HumanML3D} & \multicolumn{3}{c}{\centering AMASS-BMLrub} \\
&  &  & & Top 1 & FID & MModality & 80 & 400 & 1000 \\
\hline
1) & - & - & \attname & $0.031^{\pm.001}$ & $32.814^{\pm.176}$ & $5.293^{\pm.129}$ & 17.3 & 48.4 & 89.3 \\
2) & \checkmark & - & \attname & $0.028^{\pm.001}$ & $31.365^{\pm.171}$ & $5.714^{\pm.147}$ & 16.2 & 46.5 & 78.4 \\
3) & - & \checkmark & \attname & $0.515^{\pm.002}$ & $0.151^{\pm.002}$ & $2.214^{\pm.051}$ & 14.5 & 45.5 & 76.1 \\
4) & \checkmark & \checkmark & SAMI~\cite{zhang2024finemogen} & $0.400^{\pm.009}$ & $1.866^{\pm.009}$ & $2.983^{\pm.071}$ & 15.9 & 47.2 & 80.9 \\
5) & \checkmark & \checkmark & \attname & $0.511^{\pm.002}$ & $0.138^{\pm.002}$ & $2.426^{\pm.054}$ & 14.1 & 44.1 & 73.6 \\
\hline
\end{tabular}}
\vspace{-15pt}
\label{tab:pretrain_ablation}
\end{table}

%% file: figures/viz.tex
\begin{figure}[t]
    \centering
    \includegraphics[width=\linewidth]{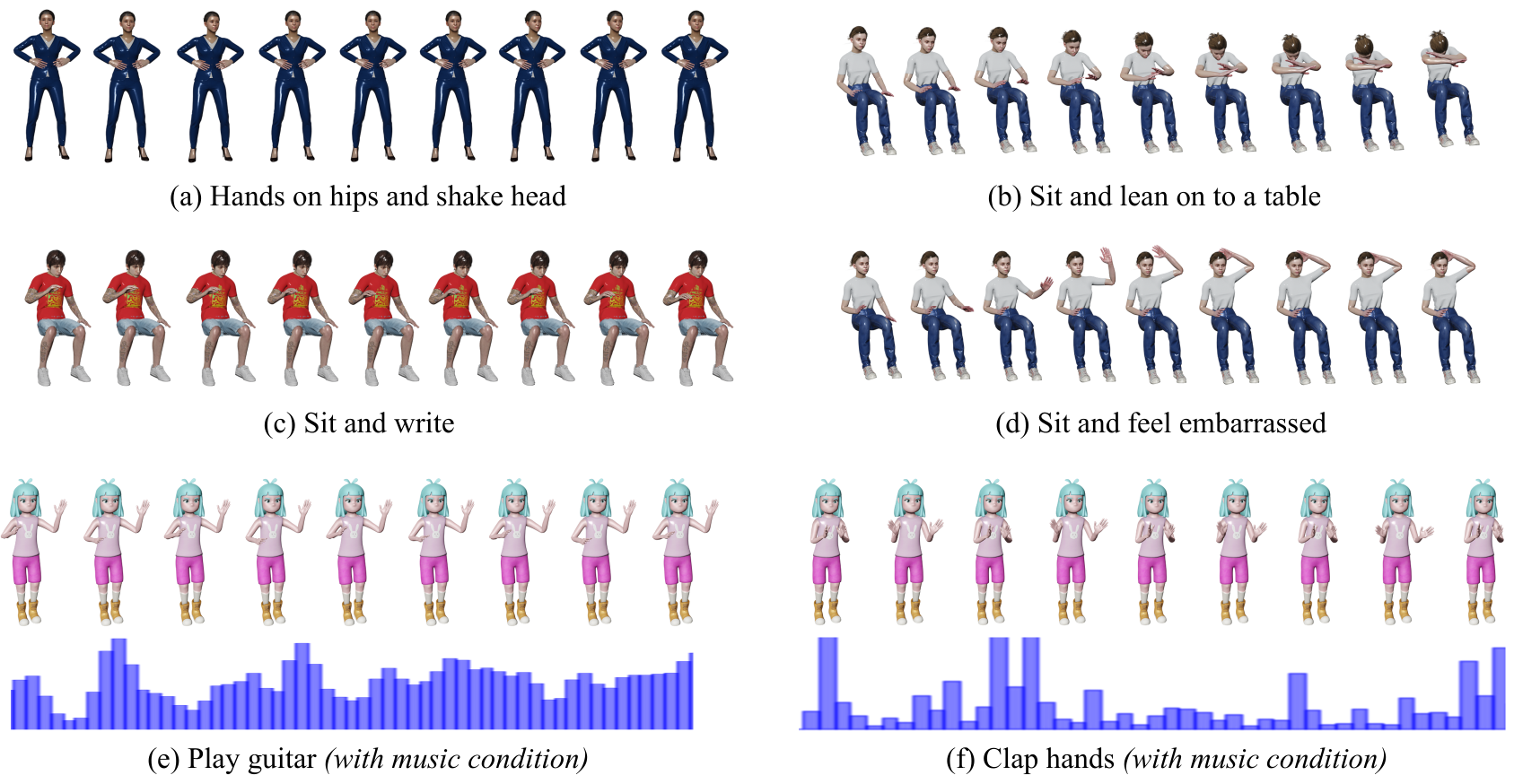}
    \caption{\textbf{Visualization results of LMM-Large.} Figure a)-d) show examples of text-driven motion generation. Figure e) and f) show synthesized motion sequences under both textual and musical constraints.}
    \label{fig:viz}
    \vspace{-10pt}
\end{figure}

%% file: secs/5_conclusion.tex
\section{Conclusion and Discussion}
In this paper, we establish a comprehensive motion-centric benchmark, \dname, comprising then conditional motion generation and motion completion tasks. We align all motion data to a unified intermediate format and convert all condition signals into token sequences that are closer in feature space. Building upon this foundation, we introduce the first large motion model, \name, capable of generating high-quality actions under multi-condition guidance. We identify and address three challenges encountered in constructing large motion models through careful model structure design, especially our used novel attention module, \attname. Our proposed \name model achieves comparable performance and even surpasses existing state-of-the-art methods. 

\noindent\textbf{Limitation.} The intermediate representation we propose can only address scenarios where entire body parts are missing, but it struggles to effectively handle cases where individual keypoints within a body part are missing. Our method of using motion translators introduces additional noise in downstream tasks, leading to a decrease in motion quality. A more flexible approach to motion representation and modeling needs to be explored and researched. Additionally, due to practical limitations in memory, our model needs to employ zero-shot methods for long-sequence motion generation, which may pose challenges for users in practical applications.

\noindent\textbf{Boarder Impact.}
The ability to generate natural human motion under flexible condition signals can highly enhance productivity. However, it may also be misused for malicious activities such as creating deceptive deepfake videos or generating realistic-looking but false evidence in legal cases.

\noindent\textbf{Acknowledegment.} This study is supported by the Ministry of Education, Singapore, under its MOE AcRF Tier 2 (MOET2EP20221- 0012), NTU NAP, and under the RIE2020 Industry Alignment Fund – Industry Collaboration Projects (IAF-ICP) Funding Initiative, as well as cash and in-kind contribution from the industry partner(s).

%% file: secs/6_supp.tex
\section{MotionVerse}

In this section, we offer additional details about the construction of \textbf{\dname} benchmark.

\subsection{Dataset Preprocess}

Based on the characteristics of each dataset, we employ different preprocessing methods. To avoid overlap between the training and test sets of different datasets, we \textbf{excluded} sequences from the training set of each benchmark that intersect with any test set motion sequences. Below, we provide detailed explanations of the processing methods for each dataset. After aligning each dataset to SMPL-X joints and processing them into TOMATO format using the same set of scripts, they are decomposed into 10 body parts. Therefore, our main focus will be on how each dataset is aligned to SMPL-X 3D joints and which body parts are included in each dataset.

\noindent\textbf{HumanML3D.} The HumanML3D dataset is annotated from two sources: AMASS data and HumanAct12. The former provides native SMPL-X annotations, while the latter offers 22 keypoint annotations based on the SMPL format. Additionally, the MotionX dataset provides facial motion data corresponding to each action sequence in HumanML3D. Therefore, there are two overall annotation formats. For data from AMASS, it includes all 10 body parts and does not require keypoint mapping. We use SMPL-X model to convert the SMPL-X beta parameters and theta parameters into 3D coordinates. For data from HumanAct12, it includes 7 body parts (excluding face, left hand, and right hand).

\noindent\textbf{KIT-ML.} We located the AMASS data corresponding to KIT-ML and utilized this portion of the AMASS data to generate the motion sequences for KIT-ML. Since there is no additional face motion data available, it comprises a total of 9 body parts.

\noindent\textbf{Motion-X.} Motion-X provides key points based on the SMPL-X format and facial expressions based on FLAME. Here, we don't need to perform additional key point conversion, and it includes all 10 body parts.

\noindent\textbf{BABEL.} BABEL is also annotated based on AMASS. Since there is no face motion available, we only consider its 9 body parts.

\noindent\textbf{UESTC.} For the UESTC dataset, we follow the processing method used in ACTOR. We use the SMPL parameters estimated from VIBE as the raw data. We use default betas parameters to obtain the 3D coordinates of each joint. Since we do not consider global orientation and global translation during evaluation, and due to the significant noise in the estimation from VIBE, we do not consider four body parts: left hand, right hand, face expression, and global configuration.

\noindent\textbf{HumanAct12.} For the HumanAct12 dataset, we employ the pre-processing method used in HumanML3D. Here, HumanAct12 does not include three body parts: left hand, right hand, and facial expression.

\noindent\textbf{NTU-RGB-D 120.} For the NTU-RGBD 120 dataset, it comes with native 3D keypoint annotations, but due to their poor accuracy, we only consider the inherent motion captured by these keypoints. Regarding the spine, we use an interpolation method to map the keypoint data from NTU-RGBD 120 to the SMPLX format for the spine. Finally, we only consider four body parts: spine, left hand, right hand, and head.

\noindent\textbf{AMASS.} AMASS provides annotations based on the SMPL-X format. We use the provided beta and theta parameters to obtain the corresponding 3D keypoints. Here, we do not consider the body part of facial expression.

\noindent\textbf{3DPW.} 3DPW provides SMPL parameters, allowing us to obtain the 3D keypoint positions in the SMPL format. Since the motion prediction task involved in 3DPW does not consider global translation, we only consider six body parts: spine, left arm, right arm, left leg, right leg, and head.

\noindent\textbf{Human3.6M.} Similar to 3DPW, we obtain keypoint sequences using SMPL parameters, which are ultimately converted into six body parts: spine, left arm, right arm, left leg, right leg, and head.

\noindent\textbf{TED-Gesture++.} TED-Gesture++ only provides keypoints for the upper body, so we consider only the spine, left arm, right arm, and head as the five body parts. For the spine, we utilize interpolation to obtain a keypoint set that conforms to SMPL-X.

\noindent\textbf{TED-Expressive.} The keypoint annotations of TED-Expressive++ are almost identical to SMPL-X. We directly selected the corresponding keypoints and removed the redundant parts. It includes all body parts except for facial expressions.

\noindent\textbf{Speech2Gesture-3D.} Similar to TED-Expressive, we directly selected the corresponding keypoints and removed the redundant parts. It includes all body parts except for facial expressions.

\noindent\textbf{BEAT.} The keypoint set of BEAT completely covers the keypoints of SMPL-X and provides facial expression, so we consider all body parts and discard the keypoints that do not exist in SMPL-X.

\noindent\textbf{AIST++.} AIST++ provides annotations based on SMPL parameters, corresponding to 7 body parts excluding face expression, left hand, and right hand.

\noindent\textbf{MPI-INF-3DHP.} Similar to 3DPW, we obtain keypoint sequences using SMPL parameters, which are ultimately converted into six body parts: spine, left arm, right arm, left leg, right leg, and head.

\subsection{Motion Translator}

During evaluation, there are three types of estimation. The first type is based on H3D vectors, primarily used in the Text2Motion datasets. For this type, we train an MLP to convert our frame representations into the corresponding representations required for evaluation. The second type is based on keypoint sequences, without considering global translation and global orientation, such as in the UESTC evaluation. Here, we also directly train an MLP for mapping. The third type considers global translation and global orientation, and is based on keypoint sequence evaluation. In this case, we first convert our representations into the keypoint sequence format and then train an MLP for mapping.

\section{Large Motion Model}

To facilitate a deeper understanding of the LMM approach, this chapter provides additional technical details.

\subsection{Diffusion Model}

This paper utilizes the Denoising Diffusion Probabilistic Model (DDPM)~\cite{ho2020denoising}, a probability generative model based on the Markov chain. 
Its essence lies in two intertwined processes: the forward diffusion process and the reverse diffusion process.

The forward diffusion process systematically injects noise into the original distribution, progressively disrupting the data's initial distribution. Starting from the original distribution $x_0\sim q(x_0)$, noise is added over $T$ steps to generate $x_1$, $x_2$,..., $x_T$. This process employs an efficient, tractable noise addition method, with Gaussian perturbation being a classic approach. The specific formula is: 
\begin{equation}
\label{eq:add-noise}
    q\left(\mathbf{x}_{t} \mid \mathbf{x}_{t-1}\right)=\mathcal{N}\left(\mathbf{x}_{t} ; \sqrt{1-\beta_{t}} \mathbf{x}_{t-1}, \beta_{t} \mathbf{I}\right), 
\end{equation}
where $\beta$ controls the amount of noise added. In the context of motion generation tasks, $x$ can be considered a series of poses. To streamline the forward process, the noise-added result at any step can be approximately calculated from $x_0$: $\mathbf{x}_{t}=\sqrt{\bar{\alpha}_{t}} \mathbf{x}_{0}+\sqrt{1-\bar{\alpha}_{t}} \boldsymbol{\epsilon}$, where $\alpha _t=1-\beta _t$ and $\bar {\alpha}_t={\textstyle \prod_{s=1}^{t}}\alpha _t $.

The reverse diffusion process is the inverse operation of adding noise, aiming to restore the original distribution from a noisy distribution. Given the difficulty and critical nature of this process, we employ deep learning models to learn the denoising process, defined as: $q(x_{t-1}\mid x_t)=M(\mathbf{x}, \text{m}, \text{c})$. During the model training phase, the supervisory objective is to minimize the difference between the predicted distribution $\hat{x}_0$ and the ground truth $x_0$.

\subsection{ArtAttention}

Following a similar approach to FineMoGen, we incorporate the temporal aspect to account for the influence of motion sequences and other condition signals across different time intervals. Specifically, we introduce the notion of time explicitly into this process. Formally, we present the following approximation for refining temporal features:
\begin{equation}
    \mathbf{Y}_{k,i} \approx \mu_{i}(x_k) = \sum\limits_{j=1}^{N_g} \mathbf{G}^{\prime}_{i,j}(x_k) \cdot \mathbf{G}^{\ast}_{i,j}(x_k)
\end{equation}
where $x_k$ represents the time position of $k$th element in the motion sequence. $\mathbf{G}^{\prime}_{i,j}(x)$ indicates the time-varied signal we derive from the feature vector $\mathbf{G}_{i,j}$ and $\mathbf{G}^{\ast}_{i,j}(x_k)$ denotes the relative significance of this template for the $k$th position. $\mathbf{G}_{i,j}$ is the $j$-th global template in the $i$-th attention head. We construct $\mathbf{G}^{\ast}_{i,j}(x_k)$ as:
\begin{equation}
    \mathbf{G}^{\ast}_{i,j}(x_k)=\frac{e^{-(x_k - \mathbf{G}^t_{i,j})^2/\sigma^2}}{\sum_{l \in [1, N_g]}e^{-(x_k - \mathbf{G}^t_{i,l})^2/\sigma^2}},
\end{equation}
In this setup, the $j$th global template of the $i$th group is considered as a set of signals propagating outward from the temporal center $\mathbf{G}^t_{i,j}$. As for $\mathbf{G}^{\prime}_{i,j}(x)$, we consider its Taylor expansion at $\mathbf{G}^t_{i,j}$:
\begin{equation}
\mathbf{G}^{\prime}_{i,j}(x)=\sum\limits_{n=0}^k\frac{\mathbf{G}^{(n)}_{i,j}}{n!}(x - \mathbf{G}^t_{i,j})^n.
\end{equation}
We use linear projections to process the original $\mathbf{G}_{i,j}$ and acquire all coefficients $\mathbf{G}^t_{i,j}, \mathbf{G}^{(n)}_{i,j},n \in [0, k]$. We perceive a global template as an anchor with its initial state defined as $\mathbf{G}^{(0)}_{i,j}$, velocity as $\mathbf{G}^{(1)}_{i,j}$, acceleration as $\mathbf{G}^{(2)}_{i,j}$ and so on. Therefore we name this method a kinetic modelling on the latent feature space. 

Moreover, to integrate influences from all signals, we adopt the square of the time difference as a metric to assess the significance of each global template. We employ a Softmax operation to standardize their weights. An immediate benefit of this modeling strategy is its flexibility in appending a new stage subsequent to the current one. This can be achieved by adjusting a bias term in $\mathbf{G}^t_{i,j}$ accordingly, facilitating our method to execute zero-shot temporal combination.

\subsection{Stylization Block}

The primary function of the stylization block is to inject the information of the timestamp $t$ into the features, thereby informing the model about the current step in the reverse process. This enhancement aids in improving the model's denoising capability. The stylization block injects information about frame rate, dataset name, and current timestep into the feature representation. Drawing inspiration from FineMoGen, we convert the timestamp $t$ into a vector $\mathbf{e_t}$. In each stylization block, $\mathbf{e_t}$ undergoes two linear transformations to generate two features $\mathbf{e_w} \in \mathbb{R}^{H \times D}$ and $\mathbf{e_b} \in \mathbb{R}^{H \times D}$. Every pose feature $\theta$ inputted into this module is optimized as $\theta^{\prime}=\theta \cdot \mathbf{e_w} + \mathbf{e_b}$, where $(\cdot)$ denotes Hadamard product.

\section{Experiments}

\subsection{Implementation Details}

\noindent\textbf{Batch Formation} Overall, we determine the sampling probability of motion sequences in each dataset based on the quality of the dataset and the diversity of the actions.

\begin{enumerate}
    \item \textbf{Text-to-Motion (40\%)}: In the task of text-to-motion, there is a wide variety of motion types, mostly consisting of high-quality motion capture data from AMASS, which is beneficial for the model. Therefore, the sampling proportion is set relatively high at 40\%. Within this category, HumanML3D provides high-quality and semantically rich motions, accounting for 15\%; Motion-X, although lower in quality, offers high diversity in motions, also at 15\%; KIT-ML and BABEL each contribute 5\%.
    \item \textbf{Unconditional Motion Generation (25\%)}: We primarily focus on the AMASS dataset, where the motion quality is generally high, aiding the model in learning motion priors. Hence, we set a relatively high proportion for this dataset.
    \item \textbf{Action-to-Motion (10\%)}: We uniformly sample sequences from the HumanAct12, UESTC, and NTU-RGBD 120 datasets.
    \item \textbf{Speech-to-Gesture (10\%)}: As BEAT is selected as the test set, we assign it half of the weight. The remaining portion is evenly distributed among TED-Gesture++, TED-Expressive, and Speech2Gesture-3D.
    \item \textbf{Music-to-Dance (5\%)}: For Music2dance, there is only one AIST++ dataset, which accounts for all the weight.
    \item \textbf{Motion Imitation (10\%)}: During training, we exclude the 3DPW dataset and only consider MPI-INF-3DHP and H36M, with both datasets equally sharing the weight.
\end{enumerate}

\begin{table}[h]
    \centering
    \resizebox{0.9\linewidth}{!}{\begin{tabular}{c|c|c|c|c}
    \hline
    \scriptsize
    \textbf{Model} & \textbf{\#Latent Dim} & \textbf{\#Layers} & \textbf{\#Experts} & \textbf{\#Params} \\
    \hline
    LMM-Tiny & 64 & 4 & 16 & 90M \\
    LMM-Small & 64 & 8 & 16 & 160M \\
    LMM-Base & 128 & 12 & 16 & 410M \\
    LMM-Large & 128 & 20 & 32 & 760M \\
    \hline
    \end{tabular}}
    \captionof{table}{Model card.}
    \label{tab:model}
\end{table}

\noindent\textbf{Model Card.} Tab.~\ref{tab:model} shows the hyperparameter of each variant.

\noindent\textbf{Mask strategy.} Considering that larger models have stronger capabilities to fit motions, to enhance the control ability of conditions, we use mask probabilities of 0.1, 0.2, 0.3, and 0.4 for LMM-Tiny, LMM-Small, LMM-Base, and LMM-Large, respectively.

\subsection{More Quantitative Results}

\begin{table}[h]
\centering
\tiny
\caption{\textbf{Quantitative results on the KIT-ML test set.}}
\label{tab:kit}
\setlength{\tabcolsep}{0.8mm}
{
\begin{tabular}{lccccccc}
\hline

\multirow{2}{1.3cm}{\centering Methods} & \multicolumn{3}{c}{\centering R Precision$\uparrow$} & \multirow{2}{0.9cm}{\centering FID$\downarrow$} & \multirow{2}{1.1cm}{\centering MM Dist$\downarrow$} & \multirow{2}{0.9cm}{\centering Diversity$\uparrow$} & \multirow{2}{0.9cm}{\centering MM$\uparrow$} \\
& Top 1 & Top 2 & Top 3 \\
\hline
Real motions & $0.424^{\pm .005}$ & $0.649^{\pm.006}$ & $0.779^{\pm.006}$ & $0.031^{\pm.004}$ & $2.788^{\pm.012}$ & $11.08^{\pm.097}$ & -\\ 
\hline

Guo \etal~\cite{guo2022generating}  & $0.370^{\pm.005}$ & $0.569^{\pm.007}$ & $0.693^{\pm.007}$ & $2.770^{\pm.109}$ & $3.401^{\pm.008}$ & $10.91^{\pm.119}$ & $1.482^{\pm.065}$ \\

T2M-GPT~\cite{zhang2023generating} &  $0.416^{\pm.006}$ &  $0.627^{\pm.006}$ & $0.745^{\pm.006}$ & $0.514^{\pm.029}$ & $3.007^{\pm.023}$ & $10.921^{\pm.108}$ & $1.570^{\pm .039}$ \\

MDM~\cite{tevet2022human}& -  & - & $0.396^{\pm.004}$ & $0.497^{\pm.021}$ & $9.191^{\pm.022}$ & $10.847^{\pm.109}$ & $\mathbf{1.907^{\pm.214}}$ \\

MotionDiffuse~\cite{zhang2024motiondiffuse} & $0.417^{\pm.004}$ & $0.621^{\pm.004}$ & $0.739^{\pm.004}$ &  $1.954^{\pm.062}$ & $2.958^{\pm.005}$ & \underline{$11.10^{\pm.143}$} & $0.730^{\pm.013}$\\

ReMoDiffuse~\cite{zhang2023remodiffuse} & $0.427^{\pm.014}$ & $0.641^{\pm.004}$ & $0.765^{\pm.055}$ & $\mathbf{0.155^{\pm.006}}$ & $2.814^{\pm.012}$ & $10.80^{\pm.105}$ & $1.239^{\pm.028}$ \\

FineMoGen~\cite{zhang2024finemogen} & \underline{$0.432^{\pm.006}$} & $0.649^{\pm.005}$ & $0.772^{\pm.006}$ & $0.178^{\pm.007}$ & $2.869^{\pm.014}$ & $10.85^{\pm.115}$ & $1.877^{\pm.093}$ \\

MoMask~\cite{guo2023momask} & $\mathbf{0.433^{\pm.007}}$ & $\mathbf{0.656^{\pm.005}}$ & $\mathbf{0.781^{\pm.005}}$ & $0.204^{\pm.011}$ & $\mathbf{2.779^{\pm.022}}$ & - & $1.131^{\pm.043}$\\
\hline
LMM-Tiny & $0.419^{\pm.018}$ & $0.627^{\pm.014}$ & $0.748^{\pm.019}$ & $0.817^{\pm.015}$ & $2.904^{\pm.022}$ & $10.85^{\pm.087}$ & $1.607^{\pm.110}$ \\
LMM-Small & $0.421^{\pm.015}$ & $0.634^{\pm.021}$ & $0.755^{\pm.017}$ & $0.471^{\pm.017}$ & $2.851^{\pm.021}$ & $10.94^{\pm.101}$ & $1.625^{\pm.114}$ \\
LMM-Base & $0.428^{\pm.015}$ & $0.648^{\pm.017}$ & $0.769^{\pm.017}$ & $0.239^{\pm.015}$ & $2.810^{\pm.018}$ & $11.05^{\pm.097}$ & $1.804^{\pm.130}$  \\
LMM-Large & $0.430^{\pm.015}$ & \underline{$0.653^{\pm.017}$} & \underline{$0.779^{\pm.014}$} & \underline{$0.137^{\pm.023}$} & \underline{$2.791^{\pm.018}$} & $\mathbf{11.24^{\pm.103}}$ & \underline{$1.885^{\pm.127}$}  \\
\hline
\end{tabular}}
\end{table}

\noindent\textbf{Text-to-Motion.} We observed that compared to its performance on HumanML3D, LMM-Large performs slightly worse on KIT-ML, which could be related to the proportion of the two datasets in batch formation. However, overall, KIT-ML also achieves accuracy comparable to the state-of-the-art, especially achieving a new state-of-the-art in terms of FID.

\begin{table}[h]
\centering
\tiny
\caption{\textbf{Quantitative results for Action-conditioned Motion Generation.} As for UESTC dataset, we report FID on the test split. MM: MultiModality.}
\setlength{\tabcolsep}{0.5mm}
{
\begin{tabular}{l|cccc|cccc}
\hline

\multirow{2}{2cm}{\centering Methods} & \multicolumn{4}{c|}{\centering HumanAct12} & \multicolumn{4}{c}{\centering UESTC} \\

& FID$\downarrow$ & Accuracy$\uparrow$ & Diversity$\rightarrow$ &  MM$\rightarrow$ & FID$\downarrow$ & Accuracy$\uparrow$ &  Diversity$\rightarrow$ &  MM$\rightarrow$ \\

\hline
Real motions & $0.020^{\pm.010}$ & $0.997^{\pm.001}$ & $6.850^{\pm.050}$ & $2.450^{\pm.040}$ & $2.79^{\pm.29}$ & $0.988^{\pm.001}$ & $33.34^{\pm.320}$ & $14.16^{\pm.06}$\\ 
\hline
Action2Motion~\cite{guo2020action2motion} & $0.338^{\pm.015}$ & $0.917^{\pm.003}$ & $6.879^{\pm.066}$ & $2.511^{\pm.023}$ & - & - & - & - \\
ACTOR~\cite{petrovich2021action} & $0.12^{\pm.00}$ & $0.955^{\pm.008}$ & $6.84^{\pm.03}$ & $2.53^{\pm.02}$ & $23.43^{\pm2.20}$ & $0.911^{\pm.003}$ & $31.96^{\pm .33}$ & $14.52^{\pm.09}$ \\
INR~\cite{cervantes2022implicit} & $0.088^{\pm.004}$ & $0.973^{\pm.001}$ & $6.881^{\pm.048}$ & $2.569^{\pm.040}$ & $15.00^{\pm.09}$ & $0.941^{\pm.001}$ & $31.59^{\pm.19}$ & $14.68^{\pm.07}$ \\
MotionDiffuse~\cite{zhang2024motiondiffuse} & \underline{$0.07^{\pm.00}$} & $\mathbf{0.992^{\pm.13}}$ & $\mathbf{6.85^{\pm.02}}$ & \underline{$2.46^{\pm.02}$} & \underline{$9.10^{\pm.437}$} & \underline{$0.950^{\pm.000}$} & \underline{$32.42^{\pm.214}$} & $14.74^{\pm.07}$ \\

\hline
LMM-Tiny & $0.105^{\pm.00}$ & $0.992^{\pm.008}$ & $6.819^{\pm.025}$ & $\mathbf{2.457^{\pm.018}}$ & $20.16^{\pm 1.78}$ & $0.917^{\pm.002}$ & $30.80^{\pm.228}$ & $\mathbf{14.29^{\pm.066}}$ \\
LMM-Small & $0.094^{\pm.00}$ & $0.963^{\pm.008}$ & $6.827^{\pm.028}$ & $2.498^{\pm.022}$ & $14.28^{\pm 1.14}$ & $0.922^{\pm.002}$ & $31.25^{\pm.231}$ & \underline{$14.42^{\pm.067}$} \\
LMM-Base & $0.087^{\pm.00}$ & \underline{$0.985^{\pm.007}$} & \underline{$6.848^{\pm.030}$} & $2.551^{\pm.022}$ & $10.36^{\pm 0.60}$ & $0.948^{\pm.000}$ & $32.39^{\pm.236}$ & $14.65^{\pm.065}$ \\
LMM-Large & $\mathbf{0.065^{\pm.00}}$ & $\mathbf{0.992^{\pm.008}}$ & $6.871^{\pm.031}$ & $2.560^{\pm.019}$ & $\mathbf{9.01^{\pm 0.54}}$ & $\mathbf{0.952^{\pm.000}}$ & $\mathbf{32.58^{\pm.254}}$ & $14.81^{\pm.064}$ \\
\hline
\end{tabular}}
\label{tab:action}
\end{table}

\noindent\textbf{Action-to-Motion.} On the action-conditioned motion generation task, each LMM-Large model achieves the best performance in terms of both FID and Accuracy. Additionally, due to exposure to more data, it exhibits higher diversity and multimodality. However, because of the nature of the action-to-motion task, an increase in both aspects does not necessarily indicate better performance.

\begin{table}[h]
\centering
\scriptsize
\caption{\textbf{Quantitative results on Speech-to-Gesture on the BEAT dataset.}}
\label{tab:speech}
\setlength{\tabcolsep}{0.8mm}
{
\begin{tabular}{l|ccc}
\hline
Methods & FGD$\downarrow$ & SRGR$\uparrow$ & BeatAlign$\uparrow$ \\
\hline

Seq2Seq~\cite{yoon2019robots} & 261.3 & 0.173 & 0.729 \\
Speech2Gesture~\cite{ginosar2019learning} & 256.7 & 0.092 & 0.751 \\
MultiContext~\cite{yoon2020speech} & 176.2 & 0.195 & 0.776 \\
Audio2Gesture~\cite{li2021audio2gestures} & 223.8 & 0.097 & 0.766 \\
CaMN~\cite{liu2022beat} & 123.7 & 0.239 & 0.783 \\
TalkShow~\cite{yi2023generating} & 91.0 & - & 0.840 \\
GestureDiffuCLIP~\cite{Ao2023GestureDiffuCLIP} & 85.17 & - & - \\
CoG~\cite{xu2023chain} & \textbf{45.87} & \textbf{0.308} & \textbf{0.931} \\

\hline
LMM-Tiny & 92.51 & 0.142 & 0.825 \\
LMM-Small & 86.94 & 0.169 & 0.836 \\
LMM-Base & 57.18 & 0.228 & 0.879\\
LMM-Large & \underline{47.95} & \underline{0.277} & \underline{0.913} \\
\hline
\end{tabular}}
\end{table}
    
\noindent\textbf{Speech-to-Gesture.} In MotionVerse, we introduce multiple speech-to-gesture datasets, and overall, LMM-Large performs impressively on the BEAT dataset as well.

\noindent\textbf{Motion Imitation} We evaluate our method on the test set of 3DPW, and obtain PA-MPJPE scores of 95.7, 91.2, 76.3, and 71.5 for LMM-Tiny, LMM-Small, LMM-Base, and LMM-Large, respectively. For reference, the PA-MPJPE scores for HMR and VIBE are 81.3 and 51.9, respectively. The performance for video-conditioning is relatively low; we will focus on addressing this issue in future work.

\begin{table}[h]
\scriptsize
\centering
\caption{\textbf{Quantitative results of motion prediction on the Human3.6M test set} for different time steps (ms). We report the MPJPE error in \textit{mm}.}
\setlength{\tabcolsep}{0.8mm}
{
\begin{tabular}{l|cccccccc}
\hline

\multirow{2}{1.5cm}{\centering Method} & \multicolumn{8}{c}{\centering Human3.6M}  \\
 & 80 & 160 & 320 & 400 & 560 & 720 & 880 & 1000 \\
\hline
siMLPe~\cite{guo2023back} & \underline{9.6} & \underline{21.7} & 46.3 & 57.3 & 75.7 & 90.1 & 101.8 & 109.4 \\
GCNext~\cite{wang2023gcnext} & \textbf{9.3} & \textbf{21.5} & 45.5 & 56.4 & \underline{74.7} & 88.9 & 100.8 & 108.7 \\
\hline
LMM-Tiny & 14.8 & 28.6 & 48.3 & 59.2 & 79.3 & 93.6 & 105.9 & 112.0 \\
LMM-Small & 14.1 & 27.4 & 47.2 & 58.1 & 78.1 & 91.5 & 103.4 & 110.3 \\
LMM-Base & 12.9 & 25.9 & \underline{44.9} & \underline{55.0} & 74.8 & \underline{87.6} & \underline{99.5} & \underline{107.1} \\
LMM-Large & 11.8 & 23.6 & \textbf{43.7} & \textbf{53.1} & \textbf{73.6} & \textbf{85.0} & \textbf{96.9} & \textbf{104.6} \\
\hline
\end{tabular}}
\vspace{-20pt}
\label{tab:h36m}
\end{table}

\noindent\textbf{Motion Prediction} Similar to the conclusion we found in 3DPW and AMASS dataset, LMM-Large performs worse than the existing work in short-term prediction and better than these work in long-term prediction.

\begin{table}[h]
\scriptsize
\centering
\caption{\textbf{Quantitative results of conditional motion completion on the HumanML3D test set}. We report the MPJPE error in \textit{mm}. We use LMM-Large for all experiments.}
\setlength{\tabcolsep}{0.8mm}
{
\begin{tabular}{cccc}
\hline

 Condition & First 25 frames & Last 25 frames & avg-MPJPE \\
\hline
No & Yes & No & 63.8 \\
No & Yes & Yes & 59.1 \\
Yes & Yes & No & 54.7 \\
Yes & Yes & Yes & 51.9 \\
\hline
\end{tabular}}
\label{tab:completion}
\end{table}

\noindent\textbf{Conditional Motion Completion} To facilitate the conditional motion completion task, we selected motion sequences from the HumanML3D test set with lengths ranging from 80 to 150 frames. We experimented with various settings and observed that the difficulty of motion inbetweening is significantly lower than motion prediction. Furthermore, introducing text conditions proved advantageous in reducing prediction errors.